\pdfoutput=1

\documentclass[11pt]{article}

\usepackage{acl}

\usepackage{times}
\usepackage{latexsym}
\usepackage[T1]{fontenc}
\usepackage[utf8]{inputenc}
\usepackage{microtype}

\usepackage{url}
\usepackage{graphics}
\usepackage{verbatim}

\title{Seeing the advantage: visually grounding word embeddings to better capture human semantic knowledge}

\author{Danny Merkx \\
  Radboud University \\ Nijmegen, The Netherlands \\
  \texttt{danny.merkx@ru.nl} \\\And
  Stefan L. Frank \\
  Radboud University \\ Nijmegen, The Netherlands \\
  \texttt{stefan.frank@ru.nl} \\\And
  Mirjam Ernestus \\
  Radboud University \\ Nijmegen, The Netherlands \\
  \texttt{mirjam.ernestus@ru.nl} \\}
  
\begin{document}

\maketitle
\begin{abstract}

Distributional semantic models capture word-level meaning that is useful in many natural language processing tasks and have even been shown to capture cognitive aspects of word meaning. The majority of these models are purely text based, even though the human sensory experience is much richer. In this paper we create visually grounded word embeddings by combining English text and images and compare them to popular text-based methods, to see if visual information allows our model to better capture cognitive aspects of word meaning. Our analysis shows that visually grounded embedding similarities are more predictive of the human reaction times in a large priming experiment than the purely text-based embeddings. The visually grounded embeddings also correlate well with human word similarity ratings. Importantly, in both experiments we show that the grounded embeddings account for a unique portion of explained variance, even when we include text-based embeddings trained on huge corpora. This shows that visual grounding allows our model to capture information that cannot be extracted using text as the only source of information.

\end{abstract}
\section{Introduction}

Distributional semantic models create word representations that quantify word meaning based on the idea that a word's meaning depends on the contexts in which the word appears. Such representations (also called embeddings) are widely used as the linguistic input for computational linguistic models, with research showing that they can account for response times in lexical decision tasks \cite{mandera,Rotaru2018,petilli}, decode brain data \cite{xu,Abnar2018}, account for brain activity during text comprehension \cite{Frank}, and correlate with human judgements of word similarity \cite{Kiela2018,derby2018,derby2020}.

While such embeddings have proven useful, they are not cognitively plausible as creating high quality embeddings requires billions of word tokens. For instance, the GloVe embeddings developed by \citet{Pennington2014} are trained on 840 billion words. It would require a human 80 years of constant reading at about 330 words per second to digest that much information. Obviously, humans are able to understand language after much less exposure, and furthermore, their sensory experience is much richer than solely reading texts. 

Embodied cognition theory poses that our conceptual knowledge is based on the entirety of our sensory experience \cite{barsalou,foglia2013}. For instance, reading the word \emph{dog} elicits sensory experiences we have with dogs, such as their sound and how they look. Embodied cognition theory thus assumes that all our sensory experiences contribute to our conceptual knowledge and processing, which should be reflected in human behaviour. Early priming studies have indeed found that visual similarities can elicit priming effects \cite{arcais,schreuder}.

If visual features are part of our conceptual knowledge, word embeddings incorporating visual features should be able to explain human behavioural data to a degree unattainable by purely text-based methods (that is, if we assume visual sensory experiences can never be fully captured by textual descriptions). That is why recent research has taken an interest in multimodal word embeddings, combining text with a second source of information, resulting in visually grounded embeddings (VGEs) in the case of visual information. 

\subsection{Related work} 

Using image tags as a source of visual context, \citet{bruni} create visual distributional semantic embeddings and use dimensionality reduction to map visual and text-based embeddings to the common VGE space. \citet{derby2018} combine text-based embeddings with the network activations of an object recognition model and show that these visual features improve the embeddings' performance in downstream tasks. \citet{petilli} use visual embeddings created by an object recognition network, and show that the embedding similarities are predictive of priming effects over and above text-based similarities.

The studies described above involve separately trained word and visual embeddings. An end-to-end approach to combine visual and linguistic information is through a deep neural network based caption-to-image retrieval (C2I) models (e.g., \citealt{Karpathy2017,Kamper2017b}). While these models are trained to encode images and corresponding written or spoken captions in a common embedding space such that relevant captions can be retrieved given an image and vice versa, the resulting embeddings have been shown to capture sentence-level semantics \cite{Chrupala2017,merkx2019NLE,merkx2021}. \citet{Kiela2018} showed that pretrained embeddings correlated better with human intuition about word meaning after being fine-tuned as learnable parameters in their C2I model.

\subsection{Current study}

In this study we investigate whether VGEs created by a C2I model explain human behavioural data. Our research question is: can VGEs capture aspects of word meaning that (current) text-based approaches cannot? To answer this question we investigate novel end-to-end trained VGEs and test them on two types of human behavioural data thought to rely on conceptual/semantic knowledge. Secondly, we take care to separate the contribution of the image modality from that of the linguistic information to see whether visual grounding captures word properties that cannot be learned by purely text-based methods. We do this by comparing our VGEs to three well-known text-based methods. 

Throughout our experiments we will use two versions of the text-based methods: custom trained on the same data as our VGEs and pretrained on large corpora. From a cognitive modelling perspective, the former of these is more interesting. While the use of large corpora may not be problematic for natural language processing applications where performance comes first, we aim to create cognitively plausible embeddings, that is, from a realistic amount of linguistic exposure. However, the inclusion of pretrained embeddings serves to answer our main research question.

\subsubsection{Semantic similarity judgements}

In our first experiment we test whether the VGEs correlate better with a measure of human intuition about word meaning than text-based embeddings. A well-known method to capture human intuition about word meaning is simply by asking subjects how similar two words are in meaning. To evaluate word embeddings, one can then see if embedding similarities for those word pairs correlate with the human judgements (e.g., \citealp{bruni,baroni2014,speer2016,Kiela2018,derby2020}). 

While the study by \citet{Kiela2018} performed a similar investigation on pretrained word embeddings fine-tuned through their C2I model, they did not take into account the fact that text might also contain visual knowledge. It is not unreasonable to assume that some visual knowledge can be gained from a large corpus of sentences solely describing visual scenes. We account for this visual knowledge from text by incorporating word embeddings trained on the image descriptions in order to investigate the contribution of the \emph{image} modality included in the VGEs.

Collecting word similarity ratings typically involves showing participants two words and asking them to rate how similar or related their meanings are, or picking the most related out of several pairs. Semantic relatedness refers to the strength of the association between two word meanings. For instance, `dog' and `leash' have a strong relationship but are not similar in meaning. Semantic similarity refers to two words sharing semantic properties, for instance `dogs' and  `cats' which are both animals that people keep as pets \cite{hill}. 

\subsubsection{Semantic priming}

In the second experiment, we test whether our VGEs are predictive of semantic priming effects from a large priming experiment \cite{hutchison}. Semantic priming effects occur when activation of a semantically related prime word facilitates the processing of the target word, resulting in shorter reaction times. If all our sensory experiences contribute to word meaning, we would expect visual perceptual properties of the prime-target pair to influence the response times.

\citet{petilli} performed a similar experiment using visual embeddings derived from activation features from an object recognition network and text-based word embeddings. Their results show that after accounting for the text-based similarity, the visual embedding similarities contribute to explaining the human reaction times only for lexical decision trails with a short stimulus onset asynchrony (SOA), and not for the naming task or long SOA trials. They attribute this to: 1) the lexical decision task being more sensitive to semantic effects than the naming task \cite{lucas}, and 2) visual information being activated in early linguistic processing and rapidly decaying \cite{pecher,schreuder}. We will further test these interactions in our own experiment.

\section{Methods}

In our experiments, we compare the VGEs from our own model with three well known text-based distributional semantic models: FastText \cite{bojanowski}, Word2Vec \cite{Mikolov2013} and GloVe \cite{Pennington2014}. For the purpose of this study, we take two approaches: 1) we train our own text-based distributional models to allow for a fair comparison to the VGEs, and 2) we use the pretrained models to investigate whether our VGEs capture semantic information that even models trained on large text corpora do not.

\subsection{Training data}

MSCOCO is a database intended for training image recognition, segmentation and captioning models \cite{Chen2015}. It has 123,287 images and 605,495 written English captions, that is, five captions paired to each image. Captions were collected by asking annotators to describe what they saw in the picture. Five thousand images (25,000 captions) are reserved as a development set.  

The captions are provided in tokenised format. In order to use them in our models we only de-capitalised all words and removed the punctuation at the end of each sentence. This results in a total of 6,184,656 word tokens and 28,415 unique word types, to which we add start- and end-of-sentence tokens for training our visually grounded model.

The images are pre-processed by resizing the images such that the shortest side is 256 pixels, while keeping the original aspect ratio. We take ten 224 by 224 crops of the image: one from each corner, one from the middle and the same five crops for the mirrored image. We use ResNet-152 \cite{He2016} pretrained on ImageNet to extract visual features from these ten crops and then average the features of the ten crops into a single vector with 2,048 features. These features are extracted by removing ResNet's classification layer and taking the activations of the penultimate layer.

\subsection{Models}

\subsubsection{Visually grounded model}

Our visually grounded model is based on the implementation by \citet{merkx2019NLE}, and we refer to that paper for the details. Here we will provide a brief overview of the model, any differences with \citet{merkx2019NLE} and the parameter settings tested in this study.

The VGE model maps images and their corresponding captions to a common embedding space. It is trained to make the embeddings for matching images and captions as similar as possible, and those for mismatched images and captions dissimilar. The model consists of two parts; an image embedder and a caption embedder. The image embedder is a single-layer linear projection on top of the image features extracted with ResNet-152. We train only the linear projection and do not further fine-tune ResNet. 

The caption embedder consists of a word embedding layer followed by a two-layer bi-directional recurrent Long Short Term Memory (LSTM) layer and finally a self-attention layer. The embedding layer has 300 dimensions and is used to represent the input words as learnable embeddings. The purpose of the LSTM is to create a contextualised hidden state for each time-step (input word). Its first layer has 1028 hidden units, while its second layer acts as a bottleneck with 300 hidden units. Finally, the purpose of the attention layer is to weigh each time-step in order to create a single fixed-length embedding for the entire caption. The attention layer has 128 hidden units. 

The image embedder has $2\times300$ dimensions so that the output matches the size of the caption embeddings. Both image and caption embedding are L2 normalised and we take their distance as the loss signal for the batch hinge loss function (see \citealp{merkx2019NLE}). The networks are trained for 32 epochs using Adam with a cyclic learning rate schedule based on \citet{Smith2017}, which varies the learning rate smoothly between $10^{-3}$ and $10^{-6}$. 

The obvious way to extract word embeddings from the trained model would be to use the trained weights of the embedding layer. Unlike for instance in GloVe, where each word's embedding is based on its full co-occurrence distribution, these embeddings are not trained specifically to capture word context or meaning and they are not necessarily the best word embeddings. However, our initial tests showed that they performed very poorly as semantic embeddings when trained from a random initialisation \footnote{\citet{Kiela2018} were able to use the input embeddings because they were initialised using pretrained embeddings.}. Rather than taking the input embeddings we create our own embeddings from the hidden representations of the model.

We create our VGEs from the hidden activations of the bottleneck LSTM layer. We use the trained caption encoder to encode all training sentences in MSCOCO. However, we remove the attention layer that creates the sentence embedding and we retain the individual activations of the LSTM at each time step. As the word representations in this layer can be used to create semantic sentence embeddings that capture human intuition about sentence meaning (as shown for instance by \citealp{merkx2019NLE,merkx2021}), we expect these representations to  better capture word meaning than the input embeddings. 

The embedding for each word is then created by summing and normalising its LSTM layer activations from all its occurrences in the dataset. As opposed to \citet{merkx2019NLE}, who used a single recurrent layer and found no further benefit of additional layers in terms of sentence embedding quality, we found that the quality of our VGEs improves when we use a two-layer LSTM, with the second layer acting as a bottleneck from which we derive the embeddings.

\subsubsection{Text-based models}

The text-based distributional models are trained on the MSCOCO captions. We train Word2Vec and FastText using the \emph{Gensim} package \cite{rehurek}. We train GloVe using the code that \citet{Pennington2014} made publicly available\footnote{\url{https://nlp.stanford.edu/projects/glove/}}.

Word2Vec and FastText were trained as the Skip-gram variant with embedding size 300, a context window of 10 and 10 negative samples. GloVe was trained with embedding size 300 and a context window of 10. All resulting word embeddings are then L2 normalised. 

In addition, we use the following pretrained vectors (all 300 dimensional): Word2Vec trained on 100 billion tokens of the Google News corpus \cite{mikolov2013b}, FastText trained on 600 billion tokens of Common Crawl \cite{mikolov2018advances} and GloVe trained on 840 billion tokens of Common Crawl \cite{Pennington2014}.
\subsection{Evaluation data}

\subsubsection{Semantic similarity judgements}

We include both semantic relatedness and similarity datasets in our analysis. It has been argued that subjects' intuitive understanding of similarity is not necessarily in line with the `scientific' notions of similarity and relatedness explained in the introduction \cite{hill}. 
Thus, if subject are not clearly instructed on these notions of similarity or relatedness, we consider the nature of the dataset undefined.

\begin{table}
    \caption{Description of the word similarity/relatedness evaluation datasets. \#available is the number of word pairs included in the evaluation. Type indicates whether the dataset captures similarity or relatedness. NA indicates subjects were not specifically instructed on the difference.}
        \resizebox{1\linewidth}{!}{
        \begin{tabular}{l | r r r}
        \hline
        \bf{Dataset} & \bf{\#word-pairs} & \bf{\#available} & \bf{type}\\ \hline
        WordSim353 & 353 & 240 & NA\\
        WordSim-S & 203 & 147 & Similarity\\
        WordSim-R & 252 & 166 & Relatedness\\
        SimLex999 & 999 & 793 & Similarity\\
        -SimLex999 Q1 & 249 & 141 & Similarity\\
        -SimLex999 Q4 & 250 & 249 & Similarity\\
        MEN & 3000 & 2889 & Relatedness\\
        RareWords & 2034 & 204 & NA\\\hline
        \end{tabular}}
    \label{datasets}
\end{table}

The WordSim353 dataset by \citet{finkelstein} contains 353 word pairs annotated with similarity ratings. While the name suggests it is a similarity rating dataset, more recent studies consider it a hybrid dataset, as subjects were not specifically instructed to judge relatedness or similarity. In a later study by \citet{agirre}, the WordSim353 data was split into similar and related pairs by annotating the word pairs. WordSim-S (similar) contains word pairs annotated as being synonyms, antonyms, identical, or hyponym-hyperonym. WordSim-R (related) contains word pairs annotated as being meronym-holonym, and pairs with none of the above relationships but with a similarity score greater than 5 (out of 10). Both sets contain all unrelated words (words not annotated with any of the above relationships and a similarity lower than 5).

SimLex999 was created with the caveats of the original WordSim353 in mind in order to create a dataset of 999 word pairs annotated for similarity rather than relatedness \cite{hill}. SimLex999 furthermore contains concreteness ratings for the word pairs. \citet{hill} divided the the dataset into concreteness quartiles based on the sum of the concreteness ratings for each pair. Using these quartiles we also look at the 25\% most concrete word pairs versus the 25\% most abstract pairs in the dataset, of course expecting our grounded model to perform best on the concrete words. 

MEN contains 3000 word pairs annotated for semantic relatedness \cite{bruni}. Ratings were collected by showing subjects two word pairs and asking them to select the most related one. MEN was specifically collected to test multi-modal models, by selecting only words that have a visual referent that appeared in a large image database. 

The RareWords dataset contains 2034 word pairs, where at least one word of each pair has a low frequency in Wikipedia \cite{luong}. Modelling low-frequency words is a challenge for many models of distributional semantics.

Not all of the words in these databases are available in our training data and thus some will not have a word embedding. Table \ref{datasets} contains an overview of the datasets described here and the number of word pairs that could be entered in our evaluations.

\subsubsection{Semantic priming}

The Semantic Priming Project (SPP) dataset \cite{hutchison} contains lexical decision times and naming times from a large priming experiment. The database is large for its kind, with 1,661 target words (and 1,661 non-words for the lexical decision task), each paired with a strong and weak prime and two unrelated primes. Furthermore, each prime-target pair was presented with a short (200ms) and a long (1200ms) SOA. Every combination of prime-target and SOA received responses from 32 subjects.

This gives us 26,576 (1661 target words $\times$ 4 priming conditions $\times$ 2 SOAs $\times$ 2 tasks) trials (disregarding the non-word word trials). We pre-processed the data by removing target words that mistakenly had more or fewer than the required four primes, trials with erroneous responses and missing data. We also lowered any capitals in the prime and target words, averaged the response times over the 32 subjects, and removed any prime-target pair that did not occur in our training data, resulting in 18,326 datapoints. 

\subsection{Analysis}

\subsubsection{Semantic similarity judgements}

To test whether the word embedding models capture human intuitions on word similarity, we use the models to calculate embedding cosine similarities for each word pair and correlate them with the human annotations. This allows us to evaluate our custom trained word embeddings to see which method best extracts word-level semantics from the MSCOCO dataset. Next, we also compute partial correlations between the human annotations and our VGE model using each of the text-based models as a control. Given that all models are trained on the same textual data, with only the VGEs having excess to the visual modality, this allows us to see whether visual grounding captures information that the text-based methods do not.

Finally we also test the partial correlations using the pretrained embeddings as a control. For each pretrained model we also add in its custom MSCOCO-trained equivalent as a control, to take into account the information that text-based models can extract from the MSCOCO captions. 

\subsubsection{Semantic priming}

Using linear regression models, we analyse how well embedding similarities predict human (log-transformed) reaction times in the SPP data using the Statsmodels package in Python \cite{seabold2010statsmodels}. We code SOA and Task as factor variables. The reaction times are not on the same scale due to differences in the required response for the lexical decision and naming tasks so we standardise the log-transformed reaction time data separately for each combination of SOA and Task. This removes the main effects of SOA and Task but we include them in the regression as we are interested in their interactions with the similarity measures.

We fit a baseline regression including the target length (number of characters), Task and SOA as regressors. We furthermore include several regressors based on SUBTLEX-US \cite{Brysbaert2009}: log-transformed word-frequency counts, contextual diversity (the number of SUBTLEX-US documents a word appears in) and the orthographic neighbourhood density (the number of SUBTLEX-US words that are one character edit away) for the target words.

Next, for each of our embedding models, we include the prime-target embedding similarities as a regressor to the baseline model. We also add two two-way interactions to test the claims made in \citet{petilli}: 1) the interaction between the embedding similarities and Task to test the difference between lexical decision and naming in terms of sensitivity to semantic effects and 2) the interaction between the embedding similarities and SOA to test their claim about the time-frame in which visual information plays a role. These regression models allow us to compare the word embedding models to each other and to the baseline using the Akaike Information Criterion (AIC), where a lower AIC indicates a better model fit. 

We also test if our VGEs can explain variance in the human reaction times that the text-based methods do not. We do this by refitting the regression models for each of the text-based similarity measures and adding the VGE similarity measures and their interactions with Task and SOA as extra regressors. For each of these regressions we then calculate the log-likelihood ratio (LLR) with the corresponding regression without the VGEs, indicating the decrease in model deviance due to adding the VGE similarity measures. Higher LLRs indicate a larger contribution of the VGEs to explaining variance in the human response times beyond what the text-based embedding similarities explain. Because the LLR follows a $\chi^2$ distribution, we can test whether including the VGEs significantly improves the regression model. 

We apply a similar approach to the pretrained text-based embeddings, but we also want to account for the information that text-based embedding models can extract from the MSCOCO captions. We do this by fitting a regression model as in the previous step except that we include both the pretrained and MSCOCO trained embeddings and their interactions with SOA and Task. We then follow the same procedure as described above by adding the VGE similarities and calculate LLRs to see if adding VGEs improves the regression fit. 

\section{Results}

\subsection{Semantic similarity judgements}

\begin{figure*}
    \centering
        \resizebox{1\linewidth}{!}{
        \includegraphics{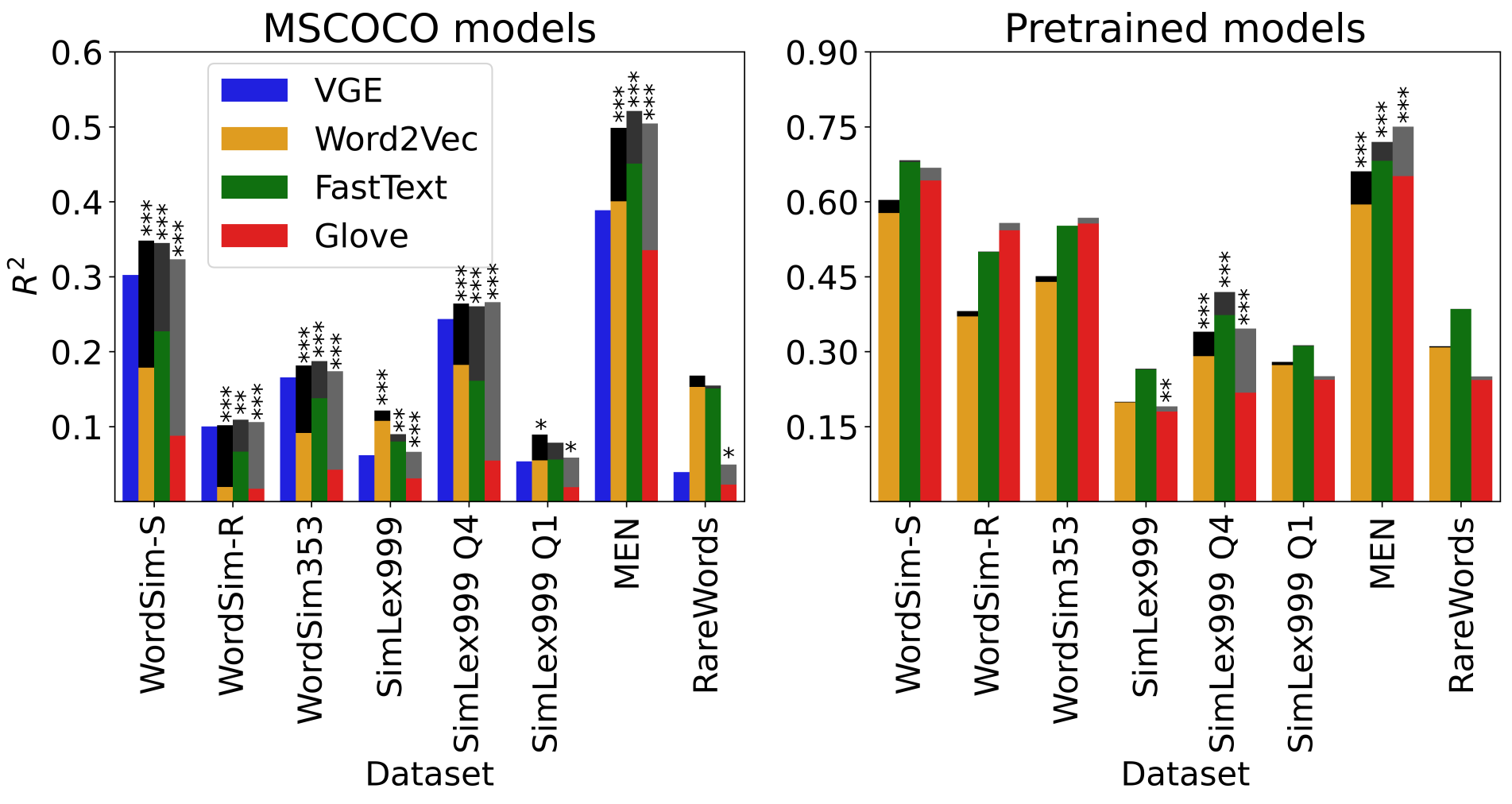}}
        \caption{The coloured bars indicate the $R^{2}$ scores of the four word embedding models. The grey-scale bars on top of the $R^{2}$ scores of the text-based models indicate the partial $R^{2}$ scores and their significance ($* p < .05, ** p < .01, *** p < .001$, corrected using the \citet{Benjamini1995} procedure with a false discovery rate of 0.05)  of the VGEs after controlling for the variance explained by that text-based model. Left panel: models trained on MSCOCO. Right panel: pretrained text-based models.}
    \label{sim_cor}
\end{figure*}

Figure \ref{sim_cor} shows the $R^{2}$ (explained variance) based on the Pearson correlation coefficients\footnote{As total explained variance is the partial $R^{2}$ plus $R^{2}$ of the control, this more clearly visualises the amount of \emph{extra} variance explained by the VGEs than Pearson's $r$.} between the human similarity annotations and the embedding similarities. All Pearson correlations were positive, as expected, except for two non-significant partial correlations which are therefore not included in the figure. 

For the MSCOCO models (left panel) we see that while GloVe has the worst performance on each dataset, there is no single best model. Furthermore, while the VGEs are outperformed by FastText and Word2Vec on SimLex999, we see that VGE performs best on the most concrete words (Q4) in SimLex999. A bit surprising then, is that VGE is outperformed by FastText and Word2Vec on MEN, which contains solely picturable nouns. 

Looking at the partial $R^{2}$, that is, the extra variance explained by the VGEs after controlling for one of the other embedding models, we see that for nearly every dataset and every model, the VGEs explain a significant portion of variance that is not explained by the text-based models. This is not very surprising on WordSim, where the VGEs were the best performing embeddings by quite a margin. However, we also see that even though the VGEs are outperformed by FastText and Word2Vec on MEN, they still explain a large extra portion of variance even though the $R^{2}$ for these models was already quite high. 

Lastly, the pretrained models (right panel) outperform the MSCOCO models. This was expected, as the used training data is several orders of magnitude larger than MSCOCO. However, here we still see that the VGEs explain a significant portion of extra variance on SimLex999 Q4 and MEN.

\subsection{Semantic priming}

\begin{table}
    \caption{AIC comparison of regression models (lower is better). $\Delta$ indicates the difference in AIC compared to the VGE model or the Baseline model. $\beta$ indicates the coefficient of the embedding similarity main effect (lower is better) and its significance.}
    \centering
        \resizebox{1\linewidth}{!}{
        \begin{tabular}{l | r r r l}
            \hline
            \bf{Model} & \bf{AIC} & \bf{$\Delta$VGE} & \bf{$\Delta$Baseline} & \bf{$\beta$}\\ \hline
            VGE & 46997.55 & --- & $-$211.04 & $-$.67***\\
            FastText & 47101.90 & 104.35 & $-$106.86 & $-$.54***\\ 
            GloVe & 47163.70 & 166.15 & $-$44.88 & $-$.20**\\ 
            Word2Vec & 47184.45 & 186.90 & $-$24.13 & $-$.22**\\
            Baseline & 47208.58 & 211.03 & --- & ---\\ \hline
        \end{tabular}}
    \label{AIC}
\end{table}

The $\Delta$AIC scores in Table \ref{AIC} show that all word embedding models trained on MSCOCO improve the regression fit above the baseline. The embedding similarity effects were all negative, that is, a higher similarity correctly predicts a lower reaction time. We furthermore see that the VGE-derived similarity measures result in the best model fit by quite a margin, as evidenced by the AIC scores and effect size. 

We also find significant interactions between Task and the embedding similarities for the VGE ($\beta = 0.201, P = 0.009$) and FastText regression models ($\beta = 0.197, P = 0.027$), meaning that the effect of embedding similarity is stronger for the lexical decision task. We find no significant interactions between the embedding similarities and SOA. 

\begin{table}
    \caption{LLRs between regression models with the indicated text-based similarity measures and the same model with the VGE similarities as extra regressors. $\beta$ VGE are the regression coefficients for the VGE similarities in each model. Higher LLRs indicate a larger improvement in model quality due to adding the VGEs.} 
    \centering
        \resizebox{1\linewidth}{!}{
        \begin{tabular}{l | l l l l}
        \hline
         & \multicolumn{2}{c}{\bf{MSCOCO}} & \multicolumn{2}{c}{\bf{+ Pretrained}} \\ 
         & LLR & $\beta$ VGE & LLR & $\beta$ VGE \\\hline
        Word2Vec & 193.72*** & $-$.77*** & 69.72*** & $-$.49***\\
        FastText & 111.46*** & $-$.63*** & 47.32*** & $-$.42***\\
        GloVe & 168.34*** & $-$.72*** & 49.80*** & $-$.36***\\\hline

        \end{tabular}}
    \label{LLR}
\end{table}

Table \ref{LLR} shows the LLRs between regression models including the (pretrained) text-based and our VGE word similarity measures and the corresponding model including only the text-based measures. We see that our VGEs significantly improve the regression fit for every type of text-based method, even when we include both the pretrained and MSCOCO text-based measures. The coefficients of the VGE effects in these models are all positive, meaning a higher VGE similarity predicts a lower reaction time. 

In the regression models including the VGEs and the MSCOCO text-based embeddings we found significant interactions between the VGE  similarities and Task in the regression models that also include Word2Vec ($\beta = 0.239, P = 0.007$) or GloVe ($\beta = 0.234, P = 0.01$) and no other interactions with Task or SOA. 

Lastly, in the regression models including the VGEs and both pretrained and MSCOCO text-based embeddings, we find significant interactions with Task for Word2Vec ($\beta = 0.312, P < 0.001$), FastText ($\beta = 0.297, P = 0.001$) and GloVe ($\beta = 0.443, P < 0.001$) vectors, and none for the VGEs. 

\section{Discussion}

We created Visually Grounded Embeddings using a caption-image retrieval model in order to test if these embeddings can capture information about word meaning that text-based approaches cannot. Importantly, by testing our VGEs on human behavioural measures typically thought to rely on conceptual/semantic knowledge, we test a central idea of embodied cognition theory, namely that our visual experiences contribute to our conceptual knowledge.

\subsection{Semantic similarity judgements}

Our first experiment showed that, when trained on the same corpus, our VGEs are on par with text-based methods. While there is no clear overall best method, the VGEs perform well on WordSim and, as might be expected, on the datasets with concrete picturable nouns. Even though the text-based methods outperform the VGEs on one of these (MEN), the VGEs still explain a significant amount of extra variance over and above what is explained by the text-based methods. This indicates that the text-based embeddings and VGEs capture non-overlapping conceptual knowledge, which we attribute to the visual grounding of the VGEs, given that the training materials were otherwise equal. 

The only database where the VGEs performed notably worse than the text-based methods was RareWords. This is perhaps because during training, the VGEs are grounded in the image corresponding to the text input, even if not all words in the sentence are visible in the picture. As the words in RareWords are generally not picturable nouns, any visual information incorporated into the word-embedding is unlikely to be helpful, or, as evidenced by the results, counterproductive.

We furthermore found that our VGEs explain additional variance in the human similarity ratings even after accounting for both the MSCOCO text-based models and pretrained models trained on massive text corpora. The fact that the VGEs explain a significant amount of extra variance even after the text-based models have seen billions of tokens of text, suggests that some aspects of word meaning cannot be captured solely from text and as well as that visual similarity plays a role in human intuition about word meaning.

\subsection{Semantic priming}

In our second experiment, the VGEs outperformed the text-based methods on explaining human reaction times from the Semantic Priming Project. Even after we account for both the MSCOCO text-based models and pretrained models in our regression, the VGEs still explain a significant amount of variance in the reaction times. 

In previous work, \citet{petilli} only found a significant contribution of visual information in the short SOA lexical decision task. We found no further proof for their hypothesis that visual information is activated in early linguistic processing and thereafter rapidly decays. Rather, we find that our VGEs improve the model quality for both short and long SOA trials. 

We did find a significant positive interaction with Task, meaning that the word embeddings explain less variance in the naming task than in the lexical decision task. This interaction was not specific to the VGEs but also occurred in the models including FastText and for all the pretrained embeddings. As claimed in \citet{petilli} and \citet{lucas} this suggests that naming tasks are in general less sensitive to semantic effects.

\section{Conclusion}

We set out to test an end-to-end approach to combining visual and textual input in a single embedding, trained on a cognitively plausible amount of data. The results from our two experiments suggest that VGEs capture aspects of word meaning that text-based approaches cannot. Even though we include word embeddings trained on corpora several orders of magnitude greater than any human's exposure to language, our VGEs still explain a unique portion of variance in both human behavioural measures. 

While our results indicate that visual grounding can provide complementary information for certain words, it may not play a role in our conceptual knowledge of rare, abstract words, as shown by our results on the RareWords corpus. Similar to \citet{petilli} this then does not support the strongest formulations of embodied cognition theory which suggest total equivalence between conceptual and sensorimotor processing \cite{glenberg}. 

Of course, one could always claim that it is just current word-embedding models that do not fully capture word meaning yet. However, given that VGEs trained on a relatively small amount of visual data can complement text-based embeddings, we do not think even larger text-corpora or more complex embedding models can ever fully capture  human semantic knowledge. The human experience is rich and varied, and our computational models can never fully capture human word knowledge while ignoring visual aspects of this experience.

\bibliographystyle{acl_natbib}
\bibliography{mybib}

\begin{thebibliography}{41}
\expandafter\ifx\csname natexlab\endcsname\relax\def\natexlab#1{#1}\fi

\bibitem[{Abnar et~al.(2018)Abnar, Ahmed, Mijnheer, and Zuidema}]{Abnar2018}
Samira Abnar, Rasyan Ahmed, Max Mijnheer, and Willem Zuidema. 2018.
\newblock {Word Embeddings have Complementary Roles in Decoding Brain
  Activity}.
\newblock \emph{Proceedings of the 8th Workshop on Cognitive Modeling and
  Computational Linguistics (CMCL 2018), Salt Lake City, Utah, USA, January 7,
  2018}, pages 57--66.

\bibitem[{Agirre et~al.(2009)Agirre, Alfonseca, Hall, Kravalova, Pasca, and
  Soroa}]{agirre}
Eneko Agirre, Enrique Alfonseca, Keith Hall, Jana Kravalova, Marius Pasca, and
  Aitor Soroa. 2009.
\newblock A study on similarity and relatedness using distributional and
  wordnet-based approaches.
\newblock In \emph{Proceedings of the 2009 Annual Conference of the North
  American Chapter of the Association for Computational Linguistics
  (HLT-NAACL-2009)}, pages 19--27, Boulder, Colorado.

\bibitem[{Baroni et~al.(2014)Baroni, Dinu, and Kruszewski}]{baroni2014}
Marco Baroni, Georgiana Dinu, and Germ{\'{a}}n Kruszewski. 2014.
\newblock Don't count, predict! a systematic comparison of context-counting vs.
  context-predicting semantic vectors.
\newblock \emph{Proceedings of the 52nd Annual Meeting of the Association for
  Computational Linguistics (Volume 1: Long Papers)}, pages 238--247.

\bibitem[{Barsalou(2008)}]{barsalou}
Lawrence~W. Barsalou. 2008.
\newblock Grounded cognition.
\newblock \emph{Annual Review of Psychology}, 59(1):617--645.

\bibitem[{Benjamini and Hochberg(1995)}]{Benjamini1995}
Yoav Benjamini and Yosef Hochberg. 1995.
\newblock Controlling the false discovery rate: a practical and powerful
  approach to multiple testing.
\newblock \emph{Journal of the Royal Statistical Society B}, 57:289--300.

\bibitem[{Bojanowski et~al.(2017)Bojanowski, Grave, Joulin, and
  Mikolov}]{bojanowski}
Piotr Bojanowski, Edouard Grave, Armand Joulin, and Tom{\'{a}}s Mikolov. 2017.
\newblock Enriching word vectors with subword information.
\newblock \emph{Transactions of the Association for Computational Linguistics},
  5:135–146.

\bibitem[{Bruni et~al.(2013)Bruni, Tran, and Baroni}]{bruni}
Elia Bruni, Namh~Khanh Tran, and Marco Baroni. 2013.
\newblock Multimodal distributional semantics.
\newblock \emph{Journal of Artificial Intelligence Research}, 49:1--47.

\bibitem[{Brysbaert and New(2009)}]{Brysbaert2009}
Marc Brysbaert and Boris New. 2009.
\newblock {Moving beyond Ku{\v{c}}era and Francis: A critical evaluation of
  current word frequency norms and the introduction of a new and improved word
  frequency measure for American English}.
\newblock \emph{Behavior Research Methods}, 41(4):977--990.

\bibitem[{Chen et~al.(2015)Chen, Fang, Lin, Vedantam, Gupta, Dollar, and
  Zitnick}]{Chen2015}
Xinlei Chen, Hao Fang, Tsung-Yi Lin, Ramakrishna Vedantam, Saurabh Gupta, Piotr
  Dollar, and C.~Lawrence Zitnick. 2015.
\newblock \href {https://doi.org/10.1093/mnras/stv1365} {{Microsoft COCO
  Captions: Data Collection and Evaluation Server}}.
\newblock \emph{arXiv preprint arXiv: 1504.00325}.

\bibitem[{Chrupa{\l}a et~al.(2017)Chrupa{\l}a, Gelderloos, and
  Alishahi}]{Chrupala2017}
Grzegorz Chrupa{\l}a, Lieke Gelderloos, and Afra Alishahi. 2017.
\newblock Representations of language in a model of visually grounded speech
  signal.
\newblock In \emph{Proceedings of the 55th Annual Meeting of the Association
  for Computational Linguistics (ACL)}, pages 613--622.

\bibitem[{Derby et~al.(2020)Derby, Miller, and Devereux}]{derby2020}
Steven Derby, Paul Miller, and Barry Devereux. 2020.
\newblock Analysing word representation from the input and output embeddings in
  neural network language models.
\newblock In \emph{Proceedings of the 24th Conference on Computational Natural
  Language Learning}, pages 442--454.

\bibitem[{Derby et~al.(2018)Derby, Miller, Murphy, and Devereux}]{derby2018}
Steven Derby, Paul Miller, Brian Murphy, and Barry Devereux. 2018.
\newblock Using sparse semantic embeddings learned from multimodal text and
  image data to model human conceptual knowledge.
\newblock In \emph{Proceedings of the 22nd Conference on Computational Natural
  Language Learning}, pages 260--270, Brussels, Belgium. Association for
  Computational Linguistics.

\bibitem[{D’Arcais et~al.(1985)D’Arcais, Schreuder, and
  Glazenborg}]{arcais}
Giovanni B.~Flores D’Arcais, Robert Schreuder, and Ge~Glazenborg. 1985.
\newblock Semantic activation during recognition of referential words.
\newblock \emph{Psychological Research}, 45(1):39--49.

\bibitem[{Finkelstein et~al.(2002)Finkelstein, Gabrilovich, Matias, Rivlin,
  Solan, Wolfman, and Ruppin}]{finkelstein}
Lev Finkelstein, Evgeniy Gabrilovich, Yossi Matias, Ehud Rivlin, Zach Solan,
  Gadi Wolfman, and Eytan Ruppin. 2002.
\newblock Placing search in context: The concept revisited.
\newblock \emph{ACM Transactions on Information Systems}, 20(1):116–131.

\bibitem[{Foglia and Wilson(2013)}]{foglia2013}
Lucia Foglia and Robert~A Wilson. 2013.
\newblock Embodied cognition.
\newblock \emph{Wiley Interdisciplinary Reviews: Cognitive Science},
  4(3):319--325.

\bibitem[{Frank and Willems(2017)}]{Frank}
Stefan~L. Frank and Roel~M. Willems. 2017.
\newblock Word predictability and semantic similarity show distinct patterns of
  brain activity during language comprehension.
\newblock \emph{Language, Cognition and Neuroscience}, 32(9):1192--1203.

\bibitem[{Glenberg(2015)}]{glenberg}
Arthur~M. Glenberg. 2015.
\newblock Few believe the world is flat: How embodiment is changing the
  scientific understanding of cognition.
\newblock \emph{Journal of Experimental Psychology}, 69(2):165--171.

\bibitem[{He et~al.(2016)He, Zhang, Ren, and Sun}]{He2016}
Kaiming He, Xiangyu Zhang, Shaoqing Ren, and Jian Sun. 2016.
\newblock Deep residual learning for image recognition.
\newblock In \emph{The IEEE Conference on Computer Vision and Pattern
  Recognition (CVPR)}.

\bibitem[{Hill et~al.(2015)Hill, Reichart, and Korhonen}]{hill}
Felix Hill, Roi Reichart, and Anna Korhonen. 2015.
\newblock Simlex-999: Evaluating semantic models with genuine similarity
  estimation.
\newblock \emph{Computational Linguistics}, 41(4):665–695.

\bibitem[{Hutchison et~al.(2013)Hutchison, Balota, Neely, Cortese,
  Cohen-Shikora, Tse, Yap, Bengson, Niemeyer, and Buchanan}]{hutchison}
Keith~A. Hutchison, David~A. Balota, James~H. Neely, Michael~J. Cortese,
  Emily~R. Cohen-Shikora, Chi-Shing Tse, Melvin~J. Yap, Jesse~J. Bengson, Dale
  Niemeyer, and Erin Buchanan. 2013.
\newblock The semantic priming project.
\newblock \emph{Behaviour Research Methods}, 45:1099--1114.

\bibitem[{Kamper et~al.(2017)Kamper, Settle, Shakhnarovich, and
  Livescu}]{Kamper2017b}
Herman Kamper, Shane Settle, Gregory Shakhnarovich, and Karen Livescu. 2017.
\newblock {Visually grounded learning of keyword prediction from untranscribed
  speech}.
\newblock \emph{{INTERSPEECH} 2017 -- 18\textsuperscript{th} Annual Conference
  of the International Speech Communication Association}, pages 3677--3681.

\bibitem[{Karpathy and Fei-Fei(2015)}]{Karpathy2017}
Andrej Karpathy and Li~Fei-Fei. 2015.
\newblock Deep visual-semantic alignments for generating image descriptions.
\newblock In \emph{The IEEE Conference on Computer Vision and Pattern
  Recognition (CVPR)}, pages 3128--3137.

\bibitem[{Kiela et~al.(2018)Kiela, Conneau, Jabri, and Nickel}]{Kiela2018}
Douwe Kiela, Alexis Conneau, Allan Jabri, and Maximilian Nickel. 2018.
\newblock {Learning visually grounded sentence representations}.
\newblock In \emph{Proceedings of NAACL-HLT 2018}, pages 408--418. Association
  for Computational Linguistics.

\bibitem[{Lucas(2000)}]{lucas}
Margery Lucas. 2000.
\newblock Semantic priming without association: A meta-analytic review.
\newblock \emph{Psychonomic Bulletin \& Review}, 7(4):618--630.

\bibitem[{Luong et~al.(2013)Luong, Socher, and Manning}]{luong}
Thang Luong, Richard Socher, and Christopher Manning. 2013.
\newblock Better word representations with recursive neural networks for
  morphology.
\newblock In \emph{Proceedings of the Seventeenth Conference on Computational
  Natural Language Learning}, pages 104--113, Sofia, Bulgaria. Association for
  Computational Linguistics.

\bibitem[{Mandera et~al.(2017)Mandera, Keuleers, and Brysbaert}]{mandera}
Pawe{\l} Mandera, Emmanuel Keuleers, and Marc Brysbaert. 2017.
\newblock Explaining human performance in psycholinguistic tasks with models of
  semantic similarity based on prediction and counting: A review and empirical
  validation.
\newblock \emph{Journal of Memory and Language}, 92:57–78.

\bibitem[{Merkx and Frank(2019)}]{merkx2019NLE}
Danny Merkx and Stefan~L. Frank. 2019.
\newblock Learning semantic sentence representations from visually grounded
  language without lexical knowledge.
\newblock \emph{Natural Language Engineering}, 25(4):451–466.

\bibitem[{Merkx et~al.(2021)Merkx, Frank, and Ernestus}]{merkx2021}
Danny Merkx, Stefan~L. Frank, and Mirjam Ernestus. 2021.
\newblock {Semantic Sentence Similarity: Size does not Always Matter}.
\newblock In \emph{{INTERSPEECH} 2021 -- 22\textsuperscript{nd} Annual
  Conference of the International Speech Communication Association}, pages
  4393--4397.

\bibitem[{Mikolov et~al.(2013{\natexlab{a}})Mikolov, Chen, Corrado, and
  Dean}]{Mikolov2013}
Tomas Mikolov, Kai Chen, Greg Corrado, and Jeffrey Dean. 2013{\natexlab{a}}.
\newblock \href {https://doi.org/10.1162/153244303322533223} {{Efficient
  Estimation of Word Representations in Vector Space}}.
\newblock \emph{arXiv preprint arXiv: 1301.3781}.

\bibitem[{Mikolov et~al.(2018)Mikolov, Grave, Bojanowski, Puhrsch, and
  Joulin}]{mikolov2018advances}
Tomas Mikolov, Edouard Grave, Piotr Bojanowski, Christian Puhrsch, and Armand
  Joulin. 2018.
\newblock Advances in pre-training distributed word representations.
\newblock In \emph{Proceedings of the International Conference on Language
  Resources and Evaluation (LREC 2018)}.

\bibitem[{Mikolov et~al.(2013{\natexlab{b}})Mikolov, Sutskever, Chen, Corrado,
  and Dean}]{mikolov2013b}
Tomas Mikolov, Ilya Sutskever, Kai Chen, Greg Corrado, and Jeffrey Dean.
  2013{\natexlab{b}}.
\newblock Distributed representations of words and phrases and their
  compositionality.
\newblock In \emph{NIPS}.

\bibitem[{Pecher et~al.(1984)Pecher, Zeelenberg, and Raaijmakers}]{pecher}
Diane Pecher, René Zeelenberg, and Jeroen G.~W. Raaijmakers. 1984.
\newblock Does pizza prime coin? perceptual priming in lexical decision and
  pronunciation.
\newblock \emph{Psychological Research}, 45(4):339--354.

\bibitem[{Pennington et~al.(2014)Pennington, Socher, and
  Manning}]{Pennington2014}
Jeffrey Pennington, Richard Socher, and Christopher~D. Manning. 2014.
\newblock Glove: Global vectors for word representation.
\newblock In \emph{Empirical Methods in Natural Language Processing (EMNLP)},
  pages 1532--1543.

\bibitem[{Petilli et~al.(2021)Petilli, Günther, Vergallito, Ciapparelli, and
  Marelli}]{petilli}
Marco~A. Petilli, Fritz Günther, Alessandra Vergallito, Marco Ciapparelli, and
  Marco Marelli. 2021.
\newblock Data-driven computational models reveal perceptual simulation in word
  processing.
\newblock \emph{Journal of Memory and Language}, 117.

\bibitem[{{\v R}eh{\r u}{\v r}ek and Sojka(2010)}]{rehurek}
Radim {\v R}eh{\r u}{\v r}ek and Petr Sojka. 2010.
\newblock {Software Framework for Topic Modelling with Large Corpora}.
\newblock In \emph{{Proceedings of the LREC 2010 Workshop on New Challenges for
  NLP Frameworks}}, pages 45--50, Valletta, Malta. ELRA.

\bibitem[{Rotaru et~al.(2018)Rotaru, Vigliocco, and Frank}]{Rotaru2018}
Armand~S. Rotaru, Gabriella Vigliocco, and Stefan~L. Frank. 2018.
\newblock {Modeling the Structure and Dynamics of Semantic Processing}.
\newblock \emph{Cognitive Science}, pages 1--28.

\bibitem[{Schreuder et~al.(1998)Schreuder, D’Arcais, and
  Glazenborg}]{schreuder}
Robert Schreuder, Giovanni B.~Flores D’Arcais, and Ge~Glazenborg. 1998.
\newblock Effects of perceptual and conceptual similarity in semantic priming.
\newblock \emph{Journal of Memory and Language}, 38(4):401--418.

\bibitem[{Seabold and Perktold(2010)}]{seabold2010statsmodels}
Skipper Seabold and Josef Perktold. 2010.
\newblock statsmodels: Econometric and statistical modeling with python.
\newblock In \emph{9th Python in Science Conference}.

\bibitem[{Smith(2017)}]{Smith2017}
Leslie~N. Smith. 2017.
\newblock Cyclical learning rates for training neural networks.
\newblock In \emph{2017 IEEE Winter Conference on Applications of Computer
  Vision (WACV)}, pages 464--472.

\bibitem[{Speer and Chin(2016)}]{speer2016}
Robert Speer and Joshua Chin. 2016.
\newblock \href {http://arxiv.org/abs/1604.01692} {{An Ensemble Method to
  Produce High-Quality Word Embeddings}}.
\newblock \emph{arXiv preprint arXiv: 1604.01692}.

\bibitem[{Xu et~al.(2016)Xu, Murphy, and Fyshe}]{xu}
Haoyan Xu, Brian Murphy, and Alona Fyshe. 2016.
\newblock Brainbench: A brain-image test suite for distributional semantic
  models.
\newblock In \emph{Proceedings of the 2016 Conference on Empirical Methods in
  Natural Language Processing}, pages 2017--2021.

\end{thebibliography}

\end{document}